\documentclass[sigconf]{acmart}
 \usepackage{amssymb}
\usepackage{algorithm}
\usepackage{algpseudocode}

\AtBeginDocument{%
  }

\begin{document}

\title{ROSfs: A User-Level File System for ROS}
\begin{abstract}
We present ROSfs, a novel user-level file system for the Robot Operating System (ROS). ROSfs interprets a robot file as a group of sub-files, with each having a distinct label. ROSfs applies a time index structure to enhance the flexible data query while the data file is under modification. It provides multi-robot systems (MRS) with prompt cross-robot data acquisition and collaboration. We implemented a ROSfs prototype and integrated it into a mainstream ROS platform. We then applied and evaluated ROSfs on real-world UAVs and data servers. Evaluation results show that compared with traditional ROS storage methods, ROSfs improves the offline query performance by up to 129x and reduces inter-robot online data query latency under a wireless network by up to 7x.
\end{abstract}






\author{Zijun Xu}

\orcid{0000-0003-3190-1584}
\affiliation{%
  \institution{ShanghaiTech University}
  \city{Shanghai}
  \country{China}
}
\email{xuzj@shanghaitech.edu.cn}

\author{Xuanjun Wen}
\affiliation{%
  \institution{ShanghaiTech University}
  \city{Shanghai}
  \country{China}
 }
\email{wenxj@shanghaitech.edu.cn}

 \author{Yanjie Song}
\affiliation{%
  \institution{ShanghaiTech University}
  \city{Shanghai}
  \country{China}
 }
\email{songyj@shanghaitech.edu.cn}

\author{Shu Yin}
\affiliation{%
  \institution{ShanghaiTech University}
  \city{Shanghai}
  \country{China}
 }
\email{yinshu@shanghaitech.edu.cn}

\renewcommand{\shortauthors}{Xu, et al.}







\maketitle
\section{Introduction} 
{Due to the availability of cost-effective sensors, better fault tolerance, and the need to tackle increasingly complex tasks, there is a growing trend of shifting from a single powerful robot to a multi-robot system (MRS) \cite{multi-auv, swarmkhaldi2015overview, mrs-GNN}. Instead of applying a gigantic all-in-one robot for diversified jobs, MRS shows its advantages in flexibility and agility in handling role switches in complex scenarios. For example, the tasks of MRS robots in an earthquake disaster should cover terrain investigation, injury detection, and salvage at once and support prompt role switch amongst them \cite{michael2014earthquake}. In this case, an MRS represents a group of heterogeneous robots with different forms (e.g., unmanned aerial and ground vehicles (UAV and UGV)), sizes, computing power (e.g., embedded edge devices and CPU+GPU platforms), and diverse sensors (e.g., LiDAR, RGB cameras). MRS broadens the development scopes of robotic algorithms and applications and provides possibilities in complex tasks with the combination of navigation \cite{arslan2016navi}, target tracking \cite{lesire2016track}, search and rescue \cite{gregory2016search}.} 


{Being the core aspect of MRS, data collaboration among multiple robots involves significant needs in data exchange and interaction. For example, upon UAV or UGV robot swarms for simultaneous localization and mapping (SLAM), one robot can assist others to locate themselves within the same terrain by sharing its constructed map, hence avoiding redundant map construction procedures \cite{mahdoui2020communicating}. Besides pre-teamed UAV or UGV robot swarms, the urban-level autonomous vehicle is another popular form of MRS, each of which is a robot node with heterogeneous architectures. MRS faces challenges in handling prompt yet ephemeral data collaborations across computing platforms. For example, cars at an intersection may exchange visual and pose status data collected from very recently to avoid collisions. How to quickly build a temporal remote data access method to target and exchange the requested data with low latency determines the efficacy of a collaborative robot network. The data collaboration becomes challenging when an MRS involves robots and data centers. Because it not only requires end-to-end data exchanging between robots but also introduces data offloading and feedback between robots and high-performance servers \cite{chinchali2021network}.} 

Robot Operating System (ROS) is a popular open-source software framework for developing robotic applications \cite{ros1}. ROS plays a critical role in implementing MRS, where the publish/subscribe model serves for real-time data communication \cite{roldan2019multi}. However, its applicability is limited in a wireless environment due to high bandwidth requirements, confining its use to the transmission of smaller messages \cite{chinchali2021network}. Furthermore, when a team of robots is executing a task, each robot needs to record the real-time data to the onboard storage to assist future data replay and analysis. An MRS can generate a huge amount of data in minutes. Researchers need to quickly collect the recorded data from onboard storage to storage servers or personal workstations for offline data analysis. The bag format provided by ROS is designed for this data logging and analysis requirement \cite{bagformat}. However, the bag format has a huge overhead when performing a file open operation, and it cannot provide efficient data query capabilities \cite{zhang2020bora}. Besides, the message data in a bag cannot be retrieved when it is open for writing, which limits the data collaboration needs when one robot requests another robot's recorded data.

\begin{figure}[hbtp]
    \centering
    \includegraphics[width=0.7\columnwidth]{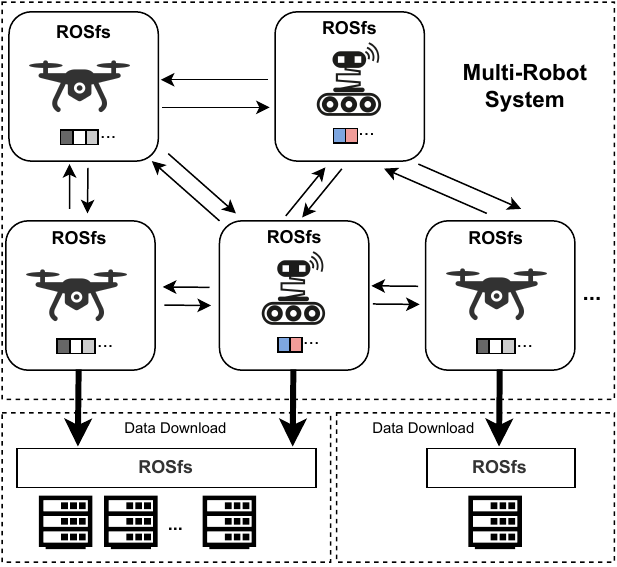}
    \caption{ROSfs Overview.}
    \label{figure:rosfs}
\end{figure}

To solve the problem of inefficient robot data collaborations, this paper proposes ROSfs, a novel user-level file system for ROS. ROSfs ensures a robot with flexible data read access to any other robot. It allows a robot to remotely read any portion of a file that is still under generation by another robot. In addition, it provides MRS with online data collaboration, allowing robots to retrieve on-disk data from any other robots within an MRS. Fig. \ref{figure:rosfs} shows the overview of ROSfs. ROSfs can be deployed on diverse types of robots to enhance efficient data acquisition and collaboration between any robots. Moreover, ROSfs also support prompt data offloading from the onboard storage of a robot to storage clusters.

The major contributions of this paper include: 
\begin{itemize}
    \item We propose ROSfs, a novel user-level file system for ROS. ROSfs is designed to enrich data collaborations in MRS.
    \item We implement a ROSfs prototype and integrate it into ROS. ROSfs can be deployed on any real-world robots installed with ROS.
    \item We perform comprehensive performance evaluation on storage servers and real UAVs under multiple data processing scenarios.
\end{itemize}

The rest of this paper is organized as follows. Section 2 provides the background and motivation for this study. Section 3 describes the detailed design and implementation of ROSfs. Section 4 presents the performance of ROSfs through multiple experiments. Section 5 summarizes the related work. Finally, Section 6 concludes the paper.

\section{Background and Motivation} 
In this section, we first introduce the background and core concepts of ROS, and then we introduce several robot data storage methods in ROS. Finally, we reveal the limitations of current storage approaches and explain the motivations for proposing ROSfs.

\subsection{Robot Operating System (ROS)}
ROS is designed as a distributed software framework that enables the developers to control the robots. The publish/subscribe model is the fundamental message-passing paradigm in ROS \cite{rospub}. Here are a few core concepts in ROS.
\begin{itemize}
    \item \textbf{Node:} In a ROS application, a node is a process that performs computation. For example, in a SLAM application, one node controls a camera, one node performs localization, one node performs path planning, and so on. 
    \item \textbf{Message:} Nodes communicate with each other by passing ROS messages. A message is a data structure comprised of typed fields like integers, floating points, strings, etc. The ROS library provides several predefined message types for commonly used sensors, like Image, Imu, PointCloud, and so on \cite{msgs}. 
    \item \textbf{Topic:} Topic is a unique name for the communication channel between publishers and subscribers. Nodes can publish messages on a given topic or subscribe to interested topics. The topic is used to identify the content of the messages. There can be multiple publishers and subscribers and they do not each other's existence.
\end{itemize}

ROS 2 is the next generation of ROS. It inherits the above concepts of ROS, but introduces a new system architecture, to support more efficient communication, higher security control, better modularity, and so on \cite{ros2}.

\subsection{Data Management of ROS}
Robot researchers need to save the onboard robot data for further data replay, analysis, and sharing \cite{rerunio}. To address this, multiple data management methods are developed or applied for ROS \cite{bagformat, sqlite, zhang2020bora, mcap}. 

\begin{figure}[hbtp]
    \centering
    \includegraphics[width=0.5\columnwidth]{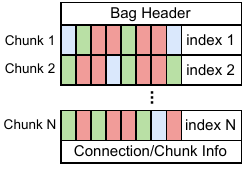}
    \caption{Bag file format.}
    \label{figure:bag-mcap}
\end{figure}

The ROS bag format \cite{bagformat} is a widely used data format in the robotic community. As shown in Fig. \ref{figure:bag-mcap}a, a bag file is made up of multiple chunks, the ROS messages are stored in these chunks, and different colors represent different topics. At the end of each chunk, there is a group of index records that saves each message's logical offset inside the current chunk. The bag file has poor data query performance on large datasets because of the huge overhead of iterating these index records upon file open \cite{zhang2020bora}. The MCAP format \cite{mcap} is the default storage format in ROS2. MCAP has a similar structure as the bag format, but it introduces more types of records like schema, and channel to support multiple data serialization formats \cite{rerunio}. Similarly, the MCAP also has an inefficient query performance as the bag format does.

BORA \cite{zhang2020bora} is a file system middleware that greatly enhances the query performance of the bag format, but it cannot be deployed on robots for real-time data recording.   SQLite \cite{sqlite} is a popular single-file SQL database engine, which is the initial storage plugin in ROS2 \cite{ros2-sqlite3}. 

\begin{figure}[hbtp]
    \centering
    \includegraphics[width=0.7\columnwidth]{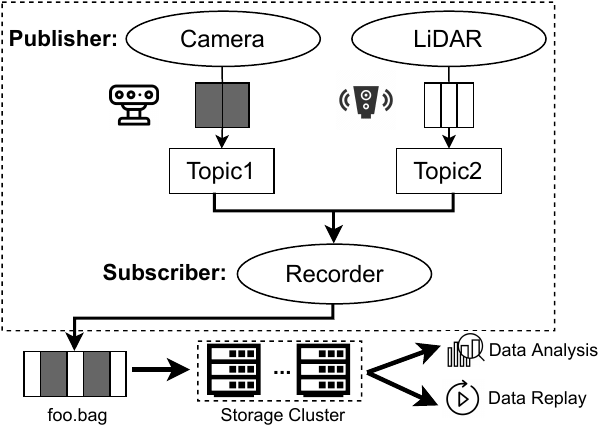}
    \caption{ROS publish/subscribe model \cite{zhang2020bora}.}
    \label{figure:pubsub}
\end{figure}

Fig. \ref{figure:pubsub} shows the process of ROS's \texttt{rosbag} tool storing messages into a bag file. The camera node and LiDAR node are publishing messages to two topics. The \texttt{rosbag} starts a recorder node to subscribe those two topics and saves the messages into a ROS bag. The bag file can be used by researchers for further data analysis. 

\subsection{Motivation}
\begin{figure}[hbtp]
    \centering
    \includegraphics[width=0.8\columnwidth]{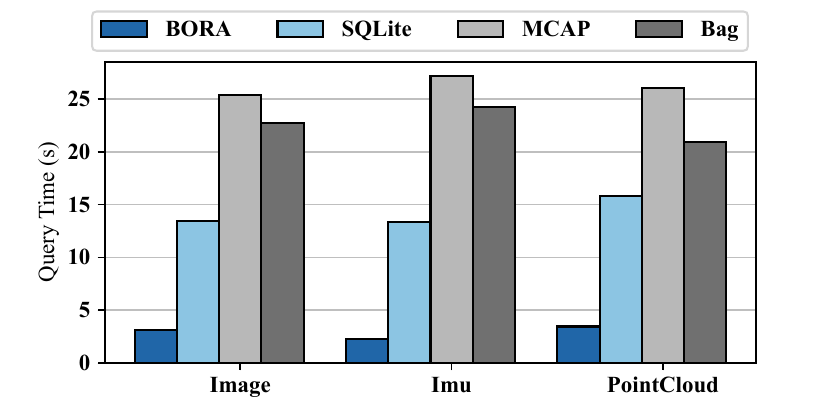}
    \caption{Query performance on a 26.5GB dataset.}
    \label{figure:large-query}
\end{figure}

\textbf{Motivation 1: }
Existing robot data storage methods has poor data query performance, which limits the data processing requirement for massive data generated by MRS. Though BORA \cite{zhang2020bora} optimizes the query performance of bag format, it cannot be used as robot data logging format. We evaluate the query performance of BORA, SQLite, MCAP, and bag by extracting three common types of ROS messages from a 26.5GB bag from the MVSEC dataset \cite{mvsec}. Table \ref{table:large-bag-topics} demonstrates the types, sizes and total number of queried messages. The test is done on a server with an Intel(R) Xeon Gold 6240L processor and a 1TB NVMe SSD. As shown in Fig. \ref{figure:large-query}, the query takes the bag format and MCAP more than 20 seconds on average, while BORA achieves up to 10x speedup compared to the bag format. This is because BORA reorganizes the data layout of the bag, thus avoiding file open overhead. However, BORA cannot be used as a real-time data logging format. Though SQLite outperforms the bag and MCAP, it still needs more than 13 seconds to perform the query operation, which is not efficient enough.  Moreover, none of these formats can support data queries during real-time data writing.

\begin{table}[htbp]
    \centering
    \caption{Selected message data from 26.5G dataset}
    \scalebox{0.8}{
    \begin{tabular}{|c|c|c|c|}
        \hline  
        \textbf{Msg Type} & \textbf{Topic} & \textbf{Size (KB)} & \textbf{Msg Number} \\\hline  
        Image      & /visensor/left/image\_raw & 352.54  & 13068  \\\hline
        Imu        & /visensor/imu             & 0.31    & 130670 \\\hline
        PointCloud & /velodyne\_point\_cloud   & 360.89  & 13068  \\\hline
    \end{tabular}
    }
    \label{table:large-bag-topics}
\end{table}

\begin{figure}[hbtp]
    \centering
    \includegraphics[width=0.9\columnwidth]{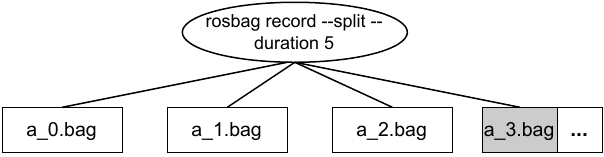}
    \caption{rosbag record with split bag files}
    \label{figure:split-bag}
\end{figure}

\textbf{Motivation 2: }
In an MRS, flexible data retrieval between robots is required, allowing any robot to access arbitrary parts of the data collected by other robots. However, the publish-subscribe communication model in ROS only provides real-time communication. Moreover, when applying the publish/subscribe model to different robots in a wireless network, data loss can happen due to the ROS's inherent code design \cite{chinchali2021network}. We also find that the original bag files cannot be queried during the writing process, as well as the MCAP file in ROS 2. However, we find a compromised approach for ROS. The \texttt{rosbag} tool provides an option to split the bag when a given duration is reached during recording. As shown in Fig. \ref{figure:split-bag}, the given duration is 5 seconds, thus each sub-bag contains 5-second message data. In this way, part of ROS data can be queried except the latest bag being written. We will compare our approach with this method in the evaluation section. 

In summary, the inefficient data query performance of the current ROS data storage formats and the lack of flexibility in data collaboration when faced with multi-node data coordination motivate us to propose ROSfs to solve these problems through efficient indexing mechanisms and redesigned data layouts.

\section{Design} 

\subsection{Design Goals}
To solve the problems of existing storage approaches in ROS, ROSfs has to consider the following issues: (1) ROSfs should provide an efficient data logging format for offline data queries when extracting interested data from a large robotic dataset. (2) ROSfs should support quick online data queries for data collaboration among multiple robots. (3) ROSfs should work with existing ROS-based systems with lower effort and cost. 

\subsection{ROSfs Architecture}

\begin{figure}[hbtp]
    \centering
    \includegraphics[width=0.65\columnwidth]{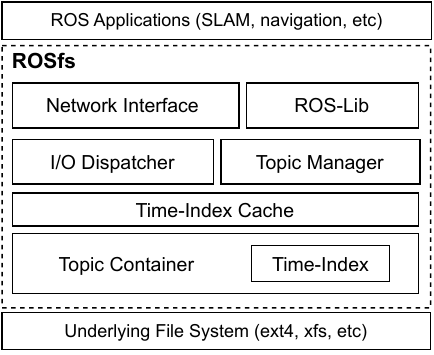}
    \caption{The architecture of ROSfs.}
    \label{figure:arch}
\end{figure}

Fig. \ref{figure:arch} illustrates the architecture of ROSfs. ROS applications can run on top of ROSfs. Network Interface provides a message-passing interface that robots can communicate with each other or with remote computing servers. I/O Dispatcher handles write and read requests from ROS and the network interface. Topic Manager maintains two hash tables, the first maps the topic name (string) and the topic ID (integer), and the second maps the topic IDs and their logical location on the underlying file system. It provides a fast lookup when dealing with different query requests. Topic Container manages the file abstraction that stores the actual message data from sensors on robots. It occupies the time-index structure as a subfile that maps message data and its offset in chronological order. ROSfs copies the time-index file into Time-Index Cache in memory upon \textit{open}.

\begin{figure}[hbtp]
    \centering
    \includegraphics[width=0.7\columnwidth]{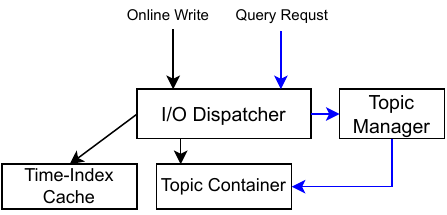}
    \caption{ROSfs IO Dispatcher.}
    \label{figure:rosfs-io}
\end{figure}

\textbf{I/O Dispatcher. } As shown in Fig. \ref{figure:rosfs-io}, the I/O dispatcher deals with two types of I/O. First, during the real-time data collection process, it captures ROS online write operations, redirecting ROS message data to the topic container structure of ROSfs. Meanwhile, it helps update the time-index cache in memory with the timestamp, and topic information from ROS messages. Second, it captures incoming query requests from local or remote robots and passes them to the topic manager. Then, the topic manager locates the positions of target data through its hash map and time-index. After ROSfs reads the target data from the topic container, it organizes the data and sends it back to the network interface.

\begin{figure}[hbtp]
    \centering
    \includegraphics[width=0.8\columnwidth]{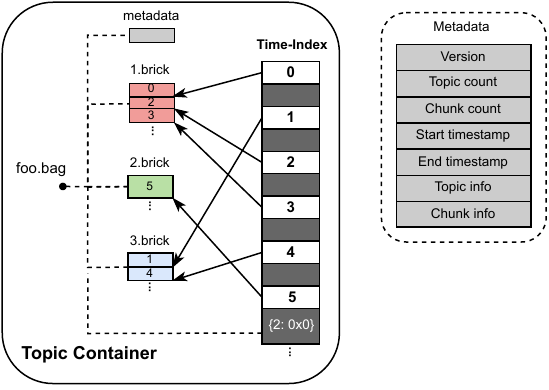}
    \caption{Internal structure of Topic Container.}
    \label{figure:rosfs-container}
\end{figure}

\textbf{Topic Container. } The topic container is the key data format of ROSfs to store ROS message data. It enables streaming data collection, fine-grained time indexing, and real-time data querying. It equips a robot with both onboard data storage and online data queries. The container employs a single-level directory structure. As depicted in Fig. \ref{figure:rosfs-container}, the topic container, named \textit{foo.bag}, houses messages from diverse ROS topics, each color-coded as a separate \textit{brick} file. The stored messages are arranged chronologically inside each brick file. Every brick file is named by its corresponding topic ID. The metadata file records the meta information of the whole container, including the topic number, chunk number, start/end timestamps, topic names, and so on. Developers can quickly get the metadata of the topic container with the help of this file. During the data collection, ROSfs appends incoming messages to the end of each brick file and updates the time-index file. 

\textbf{Time-Index. } The time-index comprises two integral components: an in-memory time-index cache and an on-disk time-index file within the topic container. The time-index cache stores temporary key-value pairs in memory which will be flushed to the on-disk time-index. The on-disk time-index within the topic container can be further accessed by clients for data queries. As shown in Fig. \ref{figure:rosfs-container}, for each ROS message, the time-index records the timestamp as the key and the respective topic ID coupled with its offset location in its corresponding brick file as value. To manage this on-disk time-index structure, we implement a B+ tree structure to persist these key-value pairs. Considering the real-time data collection circumstances of ROS, where a single write thread for each robot operates alongside potentially multiple read threads, the B+ tree can efficiently handle these simultaneous read-write demands. Furthermore, the B+ tree's inherent optimization for range queries provides an ideal solution for executing time-range queries on message timestamps. To further enhance the performance of the B+ tree, we use mmap to implement a file cache for the time-index.
  
\subsection{ROSfs Real-Time Data Write}
In the implementation of ROS, at the beginning of online data collection, the ROS recorder starts a subscriber thread that subscribes to all published ROS topics and caches the recent incoming messages into a message queue. At the same time, ROSfs creates a topic container on the local file system. ROSfs then analyzes the message popping out from the head of the queue and writes it to the corresponding brick file within the topic container.

\begin{figure}[hbtp]
    \centering
    \includegraphics[width=0.8\columnwidth]{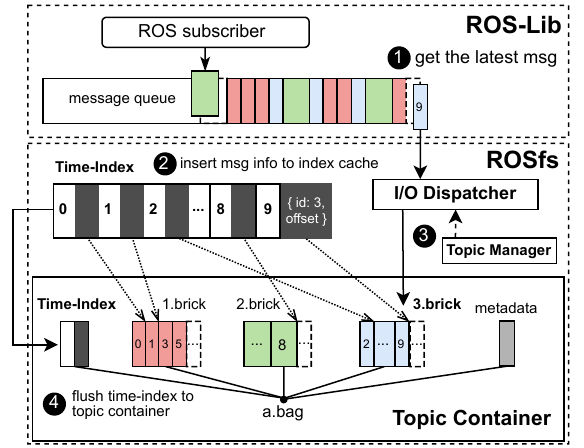}
    \caption{The real-time write process of ROSfs.}
    \label{figure:rosfs-write}
\end{figure}

Fig. \ref{figure:rosfs-write} illustrates how ROSfs writes a ROS message to the topic container of ROSfs step-by-step: (1) ROSfs gets the latest message (with timestamp 9, a.k.a msg 9) from the head of the message queue of the ROS subscriber. (2) The I/O dispatcher inserts the msg 9's info, including its timestamp, topic ID, and offset in its brick file into the time-index cache. (3) First, the topic manager performs a quick lookup in its hash table to locate the corresponding brick file path for msg 9 (its topic ID is 3). Then, the I/O dispatcher writes the message data to the end of "3.brick" file inside the topic container. (4) The in-memory time-index cache is flushed to the on-disk time-index inside the topic container by another thread. We can see that the latest message info in time-index cache is msg 9. ROSfs ensures that for any given timestamp within the time-index, its corresponding message data is readable by any client.

\subsection{ROSfs Real-Time Data Query Scenarios}
We develop three query commands in the network interface of ROSfs to access data on a remote robot: Query-Latest to read the recently collected messages, Query-History to obtain data in any historical time range, and Query-Auto to retrieve a set of data based on the fluctuating network status to meet response time requirements. Robot developers can write scripts to realize their data collaboration pattern in an MRS using the following three data query commands provided by ROSfs.

\begin{figure}[hbtp]
    \centering
    \includegraphics[width=0.8\columnwidth]{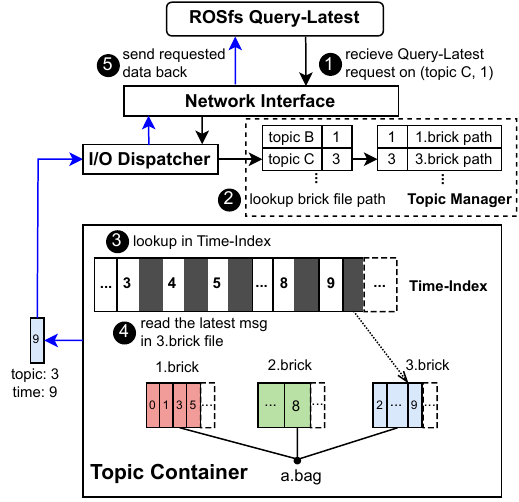}
    \caption{ROSfs Query-Latest process.}
    \label{figure:rosfs-read-q}
\end{figure}

\textbf{Query-Latest. } 
This query command enables the client robot/user to get the ``latest'' messages from any remote robot. The format of this command is \texttt{"q (topic1...topicN) time\_len"}. Fig. \ref{figure:rosfs-read-q} shows this process: (1) The robot receives the query command \texttt{\{q topicC 1\}} from a robot in the MRS, it means the other robot is requiring for the most recent 1-second topic C data from this robot (Note that each timestamp represents one second for simplicity in this figure). (2) The topic manager receives the parsed query command from the I/O dispatcher and searches its hash tables to get the target brick file path of topic C. (3) The topic manager calculates the start timestamp of the most recent 1-second is 9. Then it performs a lookup in the time-index, taking timestamp 9 as the key, and gets the offset of msg 9 in file \textit{3.brick}. (4) Having the offset, the ROSfs can directly seek to the target position in \textit{3.brick} and perform a sequential read to extract msg 9. (5) Finally, the target message is assembled by the I/O dispatcher and sent back to the network interface to be ready for passing over the network.

\textbf{Query-History. } This query command provides a time-range query operation that a robot in the MRS can require for data within a given time range from any robot as long as the data is recorded on its local ROSfs. The format of this query command is \texttt{"qh (topic1...topicN) start\_time end\_time"}. This query operation performs a similar read process as in Fig. \ref{figure:rosfs-read-q}. The detailed process of Query-History is shown in Algorithm \ref{alg:time-query}. First, the topic manager retrieves the topic IDs from the requested topic. Then, it uses the given time range and topic IDs to perform a lookup in the time-index and get a list of offsets, each offset group in the list contains a start offset and an end offset for each brick file. Finally, line 9 uses a for loop to seek to the start offset of the target brick and read the messages sequentially until reaches the end offset. 

\begin{algorithm}[h]
\caption{ROSfs Query-History}\label{alg:time-query}
\begin{algorithmic}[1]
    \Require Requested topics $topic\_list$, requested time-range $start\_time$, $end\_time$
    \Ensure ROS messages from input time range
    \State $topic\_id\_map \gets$ \{topic: topicID\} hash table in the Topic Manager
    \State $brick\_file\_map \gets$ \{topicID: brick path\} hash table in the Topic Manager
    \State $time\_index \gets$ Time-Index in the Topic Container
    \State $messages,topic\_ids \gets$ [], []
    \For {$topic$ in $topic\_list$}
    \State $topic\_ids.append(topic\_id\_map[topic])$
    \EndFor
    \State $offsets \gets time\_index.search(start\_time, end\_time, topic\_ids)$
    \For {$id$ in $topic\_ids$}
    \State $brick\_file\_path \gets brick\_file\_map[id]$
    \State $cur\_msg \gets read(brick\_file\_path, offsets[id])$ 
    \State $messages.append(cur\_msg)$
    \EndFor\\
    \Return $messages$
\end{algorithmic}
\end{algorithm}
  
\textbf{Query-Auto. }
Under a wireless network, the network bandwidth tends to fluctuate. Given the current network status, the Query-Auto command autonomously estimates the quantity of the most recent message data that can be sent within an acceptable response time. ROSfs uses a background thread to gather the real-time network status from the onboard wireless network interface and estimate the current bandwidth. Then ROSfs calculates the number of messages to be sent based on the message size of the requested topic and the estimated bandwidth. This command facilitates the clients of their data communication so that they do not need to care about the time length of data to transfer under an unstable network bandwidth. For example, when a client sends \texttt{qa topicA 1}, it means the client needs the most recent data of topicA, and it is better to receive the data in 1-second response time no matter the network status.

\subsection{Implementation of ROSfs}
The core of ROSfs is implemented in C++ with about 2000 source lines of code (SLOC) including I/O Dispatcher, Topic Manager, ROS-Lib, and the Time-Index structure. Interfaces of ROSfs such as network interface and ROS API are written in Python with 1900 SLOC. We use a cross-platform high-performance communication library called ZeroMQ \cite{zeromq} to realize the inter-robot data exchanging mechanism in MRS.

\section{Evaluation}
In this section, we first evaluate the offline query performance ROSfs on a workstation, and then we deploy it on real-world UAVs to evaluate its online query efficiency.



\subsection{ROSfs Offline Data Query Performance}
\textbf{Platform. }
In this experiment, we deploy ROSfs on a personal workstation to evaluate its offline query performance of ROSfs. The workstation represents a typical computing platform used by robotic researchers and is equipped with an Intel Core(TM) i5-9500 processor, 16GB DRAM memory, and a 1TB Samsung 980 Pro NVMe SSD. The operating system is Ubuntu 20.04.6 with Linux kernel 5.15.0, and the compiler is GCC/G++ 9.4.0. The ROS version is ROS Noetic 1.15.14. 

\textbf{Dataset. } We collect real-world ROS bags from two datasets--a RGB-D SLAM dataset \cite{sturm12iros} and a MVSEC dataset \cite{mvsec} to evaluate the offline query performance of ROSfs. Both datasets are published in the ROS bag format \cite{bagformat}. Table \ref{table:bag-organization} presents a typical data composition of a 2.8 GB handheld SLAM bag from \cite{sturm12iros}. The bag has six topics including unstructured data such as RGB image and depth image and structured data such as IMU (inertial measurement unit), camera info, and marker arrays. These topics have different message types, sizes, and publishing frequencies.

\begin{table}[htbp]
    \footnotesize
    \centering
    \caption{Data organization of a 2.8 GB ROS bag}
    \scalebox{0.9} {
\begin{tabular}{|c|c|c|c|c|}
  \hline  
  \textbf{Id}& \textbf{Topic Name}& \textbf{Message Type} & \textbf{Size} \\
  \hline  
      A& /camera/depth/image        & sensor\_msgs/Image              & 1.64 GB \\
      B& /camera/rgb/image\_color   & sensor\_msgs/Image              & 1.23 GB \\ 
      C& /imu                       & sensor\_msgs/Imu                & 8.4 MB  \\
      D& /cortex\_marker\_array     & visualization\_msgs/MarkerArray & 8.4 MB  \\
      E& /camera/depth/camera\_info & sensor\_msgs/CameraInfo         & 594 KB  \\
      F& /camera/rgb/camera\_info   & sensor\_msgs/CameraInfo         & 594 KB  \\
        \hline
\end{tabular}
}
  \label{table:bag-organization}
\end{table}

\begin{figure*}[h]
    \centering
    \includegraphics[width=2\columnwidth]{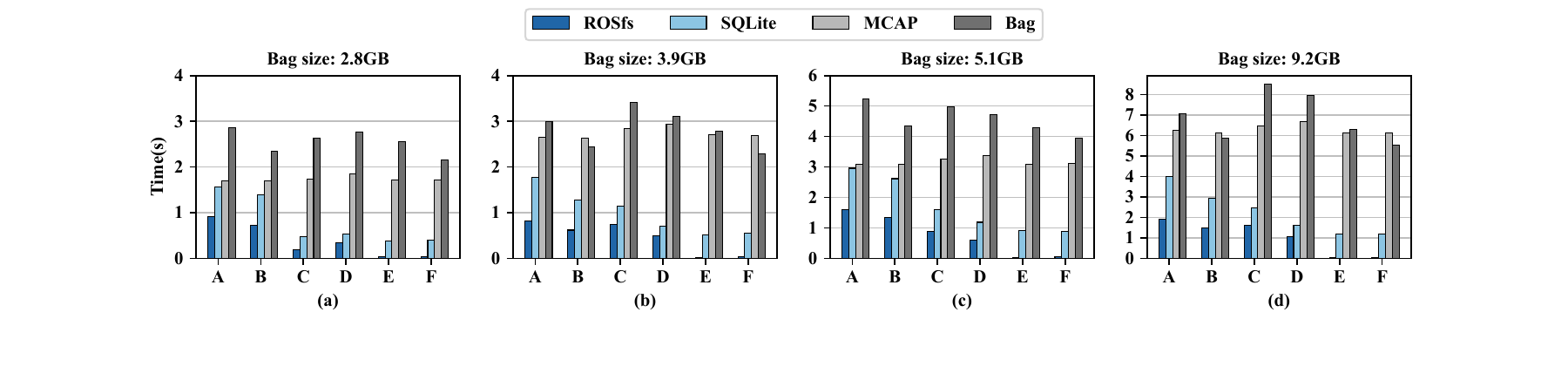}
    \caption{Query performance comparison on the specific topic for different size bags. }
    \label{figure:single-query}
\end{figure*}


\textbf{Evaluation Method. } We compare the offline query performance of ROSfs against other existing robot data storage formats, including the bag format, SQLite, and the MCAP format. The datasets are from RGB-D SLAM \cite{tum}. To convert the datasets into the ROSfs format, we replay the bags and record them using ROSfs. We use the \texttt{rosbags} tool \cite{rosbags} to convert bags into the SQLite format, and the MCAP cli tool \cite{mcap} to convert bags into the MCAP format. The bag format is provided by ROS Noetic 1.15.14, the SQLite format is provided by the rosbag2 package from ROS2 Foxy \cite{rosbag2}, and the MCAP version is v0.0.26. 

\textbf{Query by a single topic. } Fig. \ref{figure:single-query} presents the time comparisons of the query by single topic from datasets with different sizes among different storage formats. The x-axis shows six different topics listed in Table \ref{table:bag-organization}. For topic B (RGB image), as shown in Fig. \ref{figure:single-query}c, ROSfs demonstrates a performance improvement of 1.94x, 2.29x, and 3.22x over SQLite, MCAP, and Bag respectively. In particular, as shown in Fig.\ref{figure:single-query}d, ROSfs is 24.66x, 125.76x, and 128.99x faster than SQLite, MCAP, and Bag on topic E respectively. This is because, for small-size messages like topic E (about 1.3MB), the read time is negligible compared to the index building and searching time for MCAP and Bag upon file open operation. In contrast, ROSfs does not have this overhead. Though SQLite outperforms the MCAP and Bag, ROSfs still outperforms it by 57\% on average. This is because ROSfs can directly locate the target topic by brick file and extract the data taking advantage of sequential read I/O, while SQLite stores different topic data in its internal B-tree structure which requires $O(log(n))$ time on a query operation.

\begin{figure*}[h]
    \centering
    \includegraphics[width=2\columnwidth]{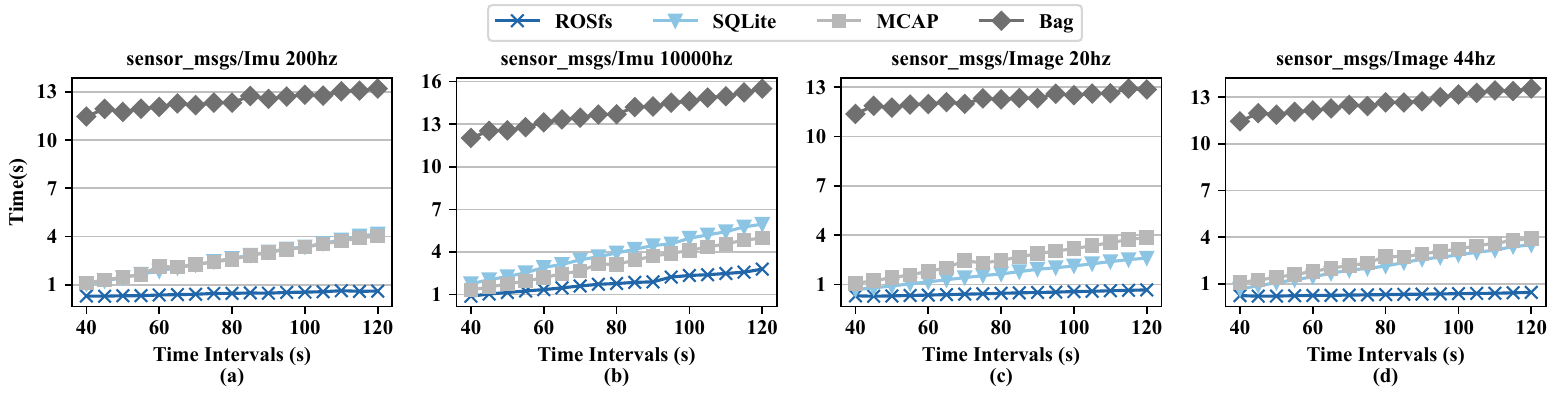}
    \caption{Time range query performance on a 26.5GB (653s) bag. }
    \label{figure:time-query}
\end{figure*}


\begin{table}[htbp]
    \centering
    \caption{Selected topics from a 26.5 GB bag, with time duration 653 seconds.}
    \scalebox{0.9} {
    \begin{tabular}{|c|c|c|}
      \hline  
      \textbf{Topic}& \textbf{Message Type} & \textbf{Frequency} \\
      \hline  
        /visensor/imu             & sensor\_msgs/Imu    &200Hz   \\
        /davis/left/imu           & sensor\_msgs/Imu    &10000Hz \\ 
        /visensor/left/image\_raw & sensor\_msgs/Image  &20Hz    \\
        /davis/left/image\_raw    & sensor\_msgs/Image  &44Hz    \\
        \hline
    \end{tabular}}
\label{table:ms-bag}
\end{table}

\textbf{Time-range query. } Time-range query is a common data query operation when extracting data within a specific time range from a large dataset. In ROS, users usually use \texttt{bag.read\_message(topics, start\_time, end\_time)} to perform a time-range query on a bag file, and ROSfs also supports this API. To evaluate its time-range query performance, we select a 26.5GB bag from the MVSEC dataset \cite{mvsec} and test on four topics (listed in Table \ref{table:ms-bag} ) of ROS messages with different publishing frequencies. 

In this experiment, we fix the start time of the query and increase the end time to increase the length of the time interval. As shown in Fig. \ref{figure:time-query}, ROSfs outperforms the ROS Bag up to 22x. In particular, when querying IMU (200hz) data, the performance improvements of ROSfs compared to SQLite, MCAP, and Bag are 6.9x, 6.7x, and 21.95x, respectively. The Bag has the worst performance because before each time-range query operation, it needs to build the index and sort the timestamps within the, which is an $O(nlogn)$ operation, where $n$ is the message number within the time range. Moreover, we observe that when the time interval increases, the query time of ROSfs has a smoother growing trend compared to the control groups. This is because ROSfs use the time-index to locate the start offset in the brick file. Thanks to the topic container structure, the time-range query on each topic is merely a sequential read operation.

\subsection{ROSfs Online Data Query Performance}
\label{eval-real-time}
\textbf{Platform. } We then deploy ROSfs on real-world UAVs and evaluate the real-time data query efficiency of ROSfs under a wireless network. UAVs are equipped with the NVIDIA\textregistered Jetson Xavier\texttrademark NX embedded computing platform, an Intel\textregistered RealSense\textregistered depth camera, and a Holybro\textregistered Pixhawk\texttrademark microcontroller. The operating system is Ubuntu 18.04.4 LTS with ROS Melodic, and the compiler is GCC/G++ 7.5.0. The UAVs are connected to a local 5Ghz WiFi network, the WiFi router is Linksys\textregistered EA7500.

\textbf{Evaluation Method. }
The binocular camera on the UAV is capable of recording topics such as RGB and depth images. To simulate a UAV performing SLAM in an unknown environment, we install the Mavros \cite{mavros} and Vins-Fusion \cite{vinsfusion} packages. During its operation, the UAV generates a total of 78 ROS topics. In the context of SLAM algorithms, image data captured by the UAV during flight and other relevant sensor data such as IMU are typically required for point cloud generation. Therefore, when evaluating performance for individual topics, we selected three topic types from Table \ref{table:drone-bag}. Considering that the binocular camera captures image data from two cameras, when evaluating the performance of querying multiple topics, we combine the depth image data from both cameras as well as the IMU data, creating the topic groups for each query request as specified in Table \ref{table:drone-topic-group}.

\begin{table}[htbp]
    \centering
    \caption{Selected Topics}
    \scalebox{0.9} {
    \begin{tabular}{|c|c|c|c|}
    \hline  
    \textbf{Topic Abbr} & \textbf{Message Type} & \textbf{Size (KB)}\\
    \hline  
    image    & sensor\_msgs/Image           & 300.07 \\\hline
    image\_c & sensor\_msgs/CompressedImage & 18.20  \\\hline
    imu      & sensor\_msgs/Imu             & 0.31   \\\hline
    \end{tabular}}
    \label{table:drone-bag}
\end{table}

\begin{table}[htbp]
    \centering
    \caption{Multiple Topic Groups}
    \begin{tabular}{|c|c|c|c|}
    \hline  
    \textbf{Name} & \textbf{Topic Combinations} & \textbf{Size (KB)} \\
    \hline  
    Group\ A & image1 + image2             & 600.1   \\ \hline
    Group\ B & image1\_c + image2\_c + imu & 36.71   \\ \hline
    \end{tabular}
    \label{table:drone-topic-group}
\end{table}

Since the ordinary ROS1 and ROS2 do not support online query patterns, we take the approach of splitting a bag or MCAP file into multiple small packets during recording to simulate ROSfs. For instance, querying the ``latest'' or historical data from the onboard robot storage while constantly appending newly collected data to a bag file. In order to investigate the impact of different splitting durations on performance, and thus analyze the maximum performance theoretically obtained by ROS1 and ROS2 by adjusting the splitting duration, we choose three typical splitting durations around our test parameters, namely 10 seconds, 2 seconds, and 0.4 seconds. Because ROS2 only supports integer splitting duration, it is impossible to perform a test for 0.4 seconds. In addition, unlike ROSfs, the bag format of ROS1 and ROS2 is incomplete before saving is completed, so if we assume a splitting duration of 10 seconds, then in the worst-case scenario, the result of querying the "latest" data is actually data from 10 seconds ago.

To perform the data collaboration between two robots, we prepare a set of ROSfs query command scripts for batch testing. For each test sub-item (topic or topic group), we generate 100 request commands with random parameters. The client robot parses all the query commands in the script and then sends these commands to the server robot at a fixed frequency (1Hz in the evaluation). Since for the same test item, the size of the returned data is theoretically identical even if the test subjects differ. Therefore, we only evaluate the server response time, which is the time from when the server receives the command from the client to the completion of the data query, and we take the average as the evaluation result. This approach helps to avoid the influence of network conditions as a variable on the observation of the results. Please note that although the test parameters are randomly generated, the request commands used for all test subjects are consistent.


\textbf{Query-Latest Performance. }
In this part, our script randomly generates the duration for Query-Latest, with 1 second and 2 seconds each accounting for half. We start the query 12 seconds after we start recording to ensure that all topics are discovered and at least one bag is saved (for ROS1 and ROS2).

\begin{figure}[h]
    \centering
    \includegraphics[width=1\columnwidth]{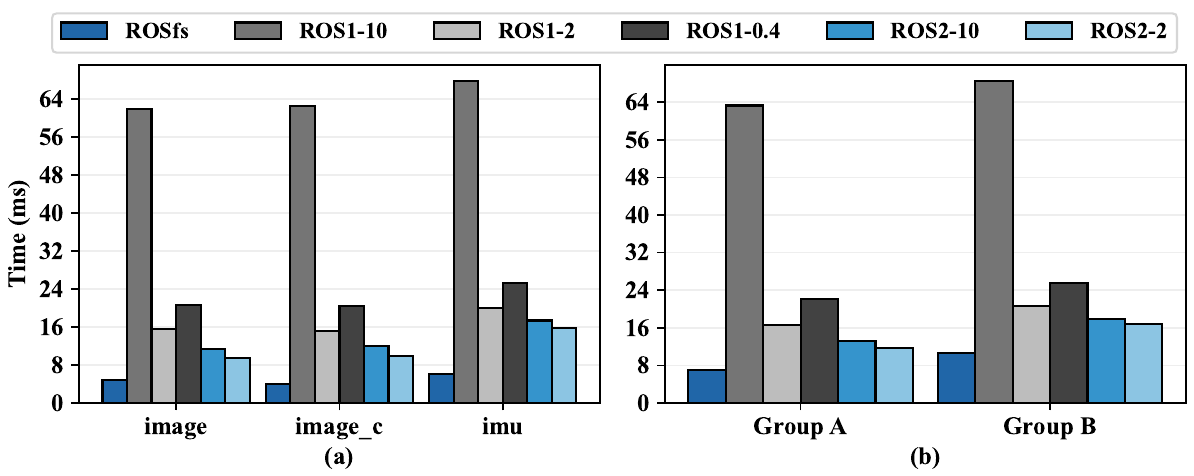}
    \caption{(a) Query-Latest (b) Query-Latest with multiple topics}
    \label{figure:eval-qbar}
\end{figure}

Fig. \ref{figure:eval-qbar} shows that ROSfs Query-Latest can finish the query of the \texttt{image} or \texttt{image\_c} topic within 4 to 5 milliseconds. Similarly, it can complete the query of the \texttt{imu} topic, which has a smaller single message size but is published more frequently, in about 6 milliseconds on average.

Under a 2-second split duration of Bag and MCAP, ROSfs is about 4x the performance of ROS1 Bag in single topic tests, and up to 1.42x higher performance compared to ROS2 MCAP. In the multiple topic tests, ROSfs is about twice as good as ROS1 Bag and outperform ROS2 MCAP by up to 69.8\%.

We found that among the three split duration settings, the simulated Query-Latest performance of ROS1 Bag and ROS2 MCAP is at best when the split duration is 2 seconds. We analyze the reason to be that the duration of our query is 1 or 2 seconds, which makes the 10-second split duration too long in terms of the query parameters, resulting in containing too many messages that are impossible to be queried, thus increasing the overhead of opening the bag. For ROS1 Bag, the performance of a 0.4-second split duration is similar to that of 2 seconds, indicating that further reducing the split duration cannot bring performance improvement, and too small a split duration will cause a large number of small files, which is unfavorable for the disk, file system, and data management.

\textbf{Query-History Performance. }
In this part, our script randomly generates time ranges with time interval lengths from 1 to 8 seconds. The time difference between the start time of the query and the start of the bag recording is a random value from 0 to 100, so we start sending requests after recording for 110 seconds to avoid requesting a time range that has not yet been recorded.

\begin{figure}[h]
    \centering
    \includegraphics[width=1\columnwidth]{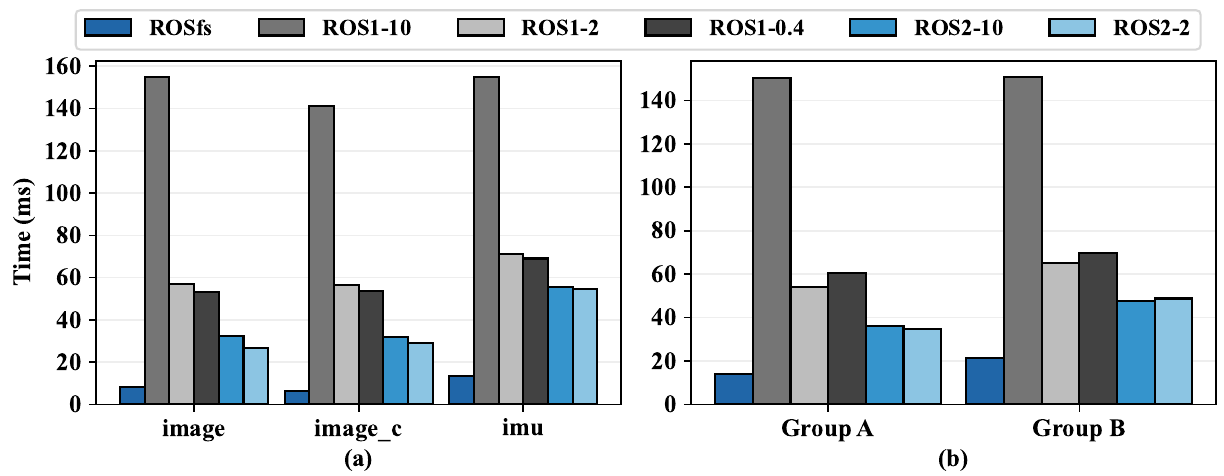}
    \caption{(a) Query-History (b) Query-History with multiple topics}
    \label{figure:eval-qtbar}
\end{figure}

According to the test results in Fig. \ref{figure:eval-qtbar} the situation for Query-History is similar to Query-Latest, but ROSfs shows a more pronounced performance advantage in this test. In the single topic test, the performance of ROSfs is approximately 7x and 4x better than the best results of ROS1 Bag and ROS2 MCAP respectively across the three split durations.

In the Query-History multi-topic test, ROSfs performs up to 2.28x better than ROS2 MCAP with a 2-second split duration, and the performance improvement further extends to more than 3x compared to ROS1 Bag. This demonstrates the advantage of ROSfs in querying on a large scale that it can efficiently completing all offset queries by Time-Index, without having to iterate the huge bag structure.

\subsection{Query-Auto Performance. }
\textbf{Evaluation Method. } We evaluate the effectiveness of the Query-Auto command in three predefined network scenarios: (1) Good, representing UAVs are close to WiFi AP with stable and good signal strength. (2) Poor: indicating the UAVs suffer from poor signal strength. We placed the UAVs behind a wall to the WiFi AP.  (3) Random: implying signal strength fluctuation as the UAVs are moving in a random route. We record all these three flying paths into ROS bags and also record the WiFi signal traces for replays. The time-elapse of the test is 100 seconds. 

\begin{figure*}[h]
    \centering
    \includegraphics[width=2\columnwidth]{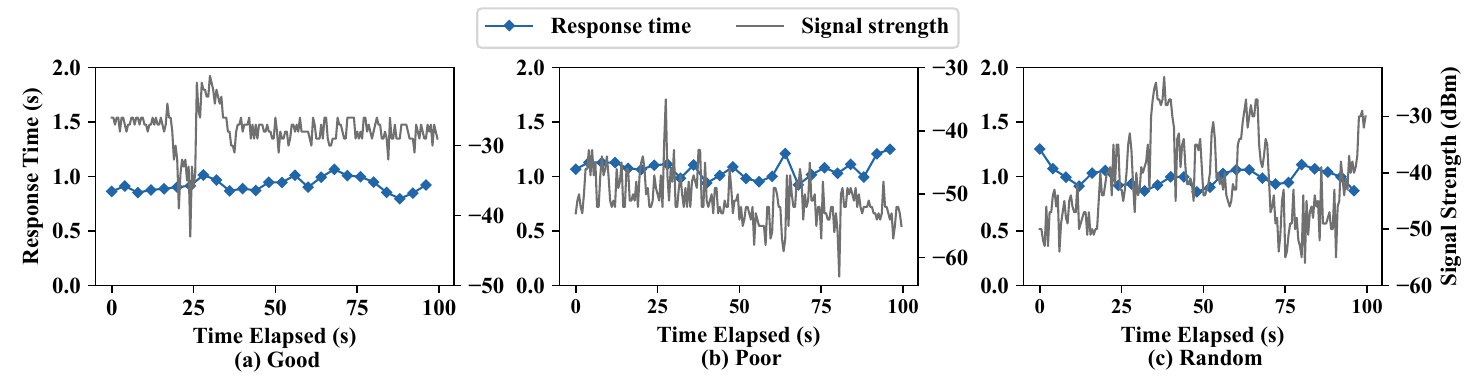}
    \caption{Query-Auto under different network conditions.}
    \label{figure:eval-qa}
\end{figure*}

\textbf{Result and Analysis. } Fig. \ref{figure:eval-qa} demonstrates the response time of each Query-Auto command. The left y-axis represents the response time, while the right y-axis shows the WiFi signal strength of the moving UAV. We observe that the response time stays around 1 second, and the response time curve has slight fluctuations compared to the signal strength curve. This experiment shows that the Query-Auto command can handle the situation when the network fluctuates.

\subsection{Concurrent Queries }
Within an MRS, a robot can receive numerous query requests from various agents concurrently. In this experiment, we evaluate the performance of ROSfs in managing concurrent multiple data query requests under a wireless network. ROS/ROS2 does not offer a communication pattern similar to ROSfs, ROS developers tend to use the publish/subscribe model to subscribe messages from another robot over the network. Moreover, ROS2 uses DDS as the transportation layer, and it supports using the QoS profile to tune its communication behavior depending on different data and network conditions \cite{ros2-qos}. 

\textbf{Control Groups. } In this experiment we choose three control groups to compare with ROSfs: (1) ROS1, the raw communication model for ROS1 based on TCPROS. (2) ROS2-best effort (ROS2-be), one of ROS2's predefined QoS profiles, designed for transmitting sensor data like RGB images, IMU data, etc. It applies the ``Best effort'' option for ``reliability'', but may lose messages if the network is not robust. (3) ROS2-default, the default QoS profile in ROS2, it applies the ``Reliable'' option for ``reliability'', it may retries if message transmission fails.

\textbf{Evaluation Method. } On the server side, one UAV is publishing messages in a fixed frequency, we select the ``image'' message in Table \ref{table:drone-bag} with a 30Hz publishing rate. On the client side, we start several data query threads using the Query-Latest to retrieve data from the target UAV. We apply the same query commands as ROSfs to the control groups. For ROS1/ROS2, they need to start a subscriber node to get data, we keep them subscribing to the target topic for 1 second and then finish subscribing. We first measure the average receive rate of all the requesting threads. It is calculated by \textit{the number of received messages / 30}, since the publishing rate is 30Hz.  We then measure the average latency of each thread by calculating the elapsed time between the time when the first piece of message arrives and the time the thread sends the query command. Note that for the thread that fails to get data, we add a 2-second penalty to its latency, because it takes a thread about 2 seconds to restart a subscriber to get the messages again.

\begin{figure}[h]
    \centering
    \includegraphics[width=1\columnwidth]{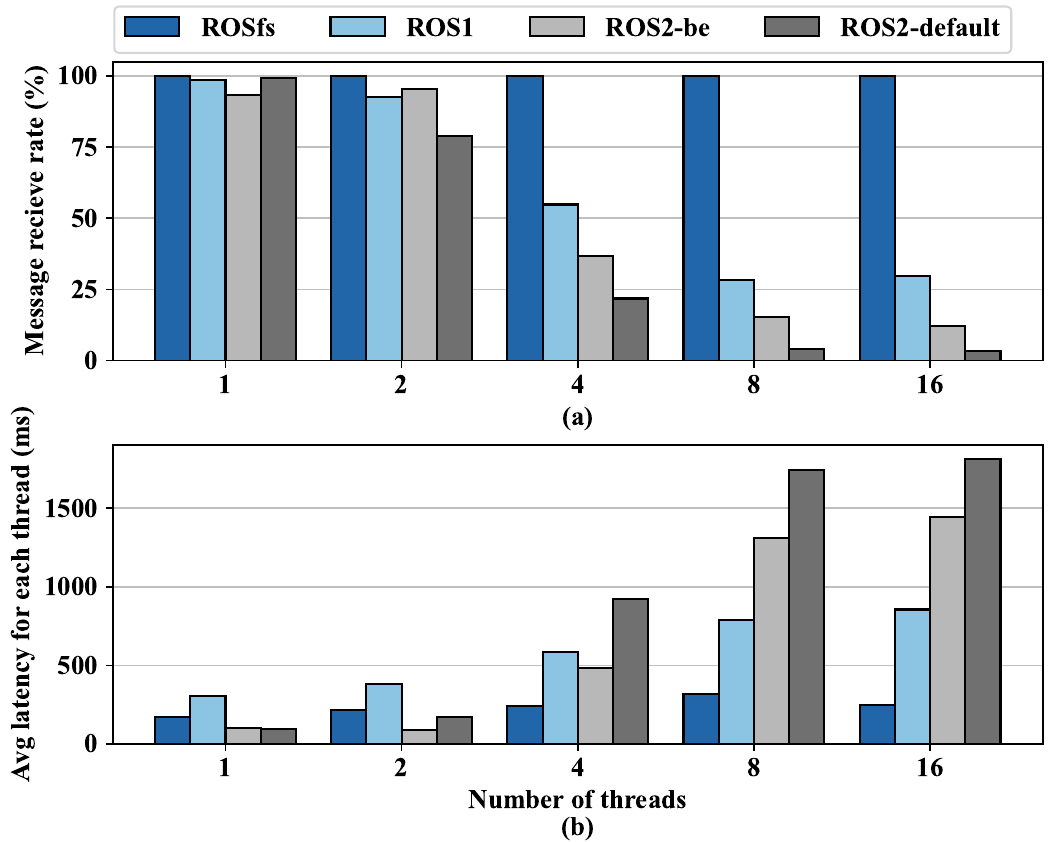}
    \caption{Multi-thread Concurrent Online Data Query}
    \label{figure:multi-read}
\end{figure}

\textbf{Result and Analysis. }As shown in Fig. \ref{figure:multi-read}a, the x-axis represents the number of threads requesting data, and the y-axis presents the average message receive rate of threads when handling concurrent data requests. We can observe that ROSfs can guarantee a 100\% message reception rate when handling simultaneous query requests compared to the control group. In contrast, when the number of threads exceeds 2, the message receive rates of ROS1 and ROS2 decrease severely. This is because ROSfs do not use the publish/subscribe model, thus having a lower requirement for real-time bandwidth, allowing it to handle multiple concurrent requests simultaneously. In particular, ROS2-default only got 4.05\% and 3.42\% messages received when the thread number is 8 and 16 respectively, this is because it uses the ``reliable'' setting in the QoS profile and tends to retry when network bandwidth is not enough. 

Fig. \ref{figure:multi-read}b demonstrates the average latency for the threads to receive the first message. We can observe that the latency of ROSfs increases gradually with the number of threads. ROSfs achieves performance improvements 2.2x, 1.8x, and 3.4x compared to ROS1, ROS2-be and ROS2-default respectively, when thread number is 4. ROSfs also outperforms the control groups when thread number $\geq$ 4. ROS2 has a relatively lower latency when the number of threads is 1 or 2 compared to ROS1, this is because the DDS and rcl layer of ROS2 have lower overhead when starting a subscriber node. However, when the number of threads exceeds 4, the latency of ROS2 increases significantly, which is because many threads in ROS2 cannot subscribe messages in time, resulting in 2-second penalty. These results demonstrate that ROSfs is much more reliable and efficient than ROS1/ROS2 to when dealing with concurrent data query requests.

\section{Related Work}

\textbf{Robot Data Storage Formats. } A number of data storage formats are developed to help researchers to analyze the robotic data. The bag format \cite{bagformat} originated in ROS1 is the most widely used file format to record and share robotic datasets. BORA \cite{zhang2020bora} is a file system middleware that optimizes the bag format for better query performance, but lacks the capability of real-time data logging. The MCAP format \cite{mcap} is default data logging format in ROS2. However, it adopts a similar structure the bag format, thus leading to inefficient query performance. 

\textbf{Database Systems}
Many database systems have been applied to facilitate robotic data requirements. SQLite \cite{sqlite} is a widely used single-file SQL database engine. SQLtie is the default logging format in the early versions of ROS2 \cite{ros2-sqlite3}. SLAMinDB \cite{fourie2017slamindb} implements a centralized server using one Neo4J (a graph database) database \cite{neo4j} and one MongoDB \cite{MongoDB} database to store different robot states for SLAM applications. The NoSQL database Cassandra was integrated into ROS to store robotic data in a smart environment \cite{dietrich2014ros}. However, these databases are not designed for real-time data query between multi-robots. It is also hard for users to port these databases to existing ROS based system. \cite{zhang2020bora}.

Table \ref{table:rosfs-comp} presents a comparison between ROSfs and other robot data storage solutions. We can observe that except for ROSfs, none of the solutions can support real-time data retrieval when file being written, thus limiting the demand for data collaboration in MRS. ROSfs has all the features including real-time data collection, fast offline data query and flexible real-time query. 



\begin{table}[h]
    \centering
    \caption{ROS comparison with other storage methods}
    \scalebox{0.9}{
    \begin{tabular}{|c| c | c | c |}
    \hline  
        & \textbf{Data collection} & \textbf{Offline query} & \textbf{Real-time query}  \\ \hline 
        Bag     & $\checkmark$  & $\checkmark$(slow) & $\times$ \\ \hline
        MCAP    & $\checkmark$  & \centering$\checkmark$(slow) & $\times$ \\\hline 
        SQLite  & $\checkmark$  & $\checkmark$(fine) & $\times$ \\ \hline
        BORA    & $\times$      & $\checkmark$(fast) & $\times$ \\ \hline
        ROSfs   & $\checkmark$  & $\checkmark$(fast) & $\checkmark$ \\ \hline
    \end{tabular}
    }
    \label{table:rosfs-comp}
\end{table}


\section{Conclusions}
In this paper, we present ROSfs, a user-level file system for ROS, providing robot data management and collaboration for multi-robot systems (MRS). ROSfs provides an efficient onboard data storage format that optimizes data query performance. Facilitated by this format, ROSfs allows data query operations in real-time data collection. Hence, with ROSfs any robot in an MRS can exchange on-disk data with other robots or remote servers, which extends the data collaboration scenarios of MRS applications. We implement ROSfs and integrate it into ROS, then evaluate its performance and effectiveness. Our results demonstrate that ROSfs significantly improves the data query performance by up to 129x compared to existing robot data storage formats. Moreover, ROSfs effectively supports online data collaboration among multiple robots under a wireless network. We anticipate ROSfs can be applied in real-world MRS to accomplish various tasks.


\bibliographystyle{ACM-Reference-Format}
\bibliography{sample-sigconf}

\end{document}